\documentclass{article}
\usepackage{spconf,amsmath,graphicx}
\usepackage{mathrsfs}
\usepackage{float}
\usepackage{amssymb}
\usepackage{cite}
\usepackage{booktabs}
\usepackage{multicol,listings,multirow}
\usepackage{lipsum}
\usepackage{tikz}
\usepackage{amssymb}
\usepackage{wrapfig}
\usepackage{array}
\usepackage{makecell}
\usepackage{booktabs}
\usepackage{array}
\usepackage{color}
\usepackage{makecell}
\usepackage[colorlinks=true,
            linkcolor=blue,
            filecolor=blue,      
            urlcolor=blue,       
            citecolor=blue
            ]{hyperref}
\usepackage[marginal]{footmisc}
\usepackage[misc]{ifsym}
\usepackage{soul}
\usepackage{enumitem}
\setlist{nosep, leftmargin=14pt}

\usepackage{mwe} 

\title{UNO-QA: An Unsupervised Anomaly-Aware Framework with Test-Time Clustering for OCTA Image Quality Assessment}
\name{Juntao Chen$^{1}$, Li Lin$^{1,2}$, Pujin Cheng$^{1}$, Yijin Huang$^{1,3}$, Xiaoying Tang$^{1,4*}$
\thanks{* {Corresponding author is Dr. Xiaoying Tang  (\href{mailto:tangxy@sustech.edu.cn}{tangxy@sustech.edu.cn})}.}}
\address{$^1$Department of Electronic and Electrical Engineering, Southern University of Science and Technology,\\ Shenzhen, China\\
$^2$Department of Electrical and Electronic Engineering, The University of Hong Kong,\\Hong Kong SAR, China\\
$^3$School of Biomedical Engineering, The University of British Columbia,\\ Vancouver, Canada\\
$^4$Jiaxing Research Institute, Southern University of Science and Technology,\\ Jiaxing, China}
\begin{document}
%
\maketitle
\begin{abstract}
Medical image quality assessment (MIQA) is a vital prerequisite in various medical image analysis applications. Most existing MIQA algorithms are fully supervised that request a large amount of annotated data. However, annotating medical images is time-consuming and labor-intensive. In this paper, we propose an unsupervised anomaly-aware framework with test-time clustering for optical coherence tomography angiography (OCTA) image quality assessment in a setting wherein only a set of high-quality samples are accessible in the training phase. Specifically, a feature-embedding-based low-quality representation module is proposed to quantify the quality of OCTA images and then to discriminate between outstanding quality and non-outstanding quality. Within the non-outstanding quality class, to further distinguish gradable images from ungradable ones, we perform dimension reduction and clustering of multi-scale image features extracted by the trained OCTA quality representation network. Extensive experiments are conducted on one publicly accessible dataset sOCTA-3×3-10k, with superiority of our proposed framework being successfully established.

\end{abstract}
\vspace{-0.4cm}
\section{Introduction}
\label{sec:intro}
\vspace{-0.2cm}
Optical coherence tomography angiography (OCTA) is a novel non-invasive imaging method and is being widely used in clinical diagnoses \cite{lin2021bsda,peng2021fargo}. It employs the motion contrast imaging technology to get high-resolution volumetric blood flow data and produce angiographic images \cite{de2015review}. \begin{figure}[htbp]
	\centering
	\vspace{-0.2cm}
	\setlength{\abovecaptionskip}{-0.2cm}   
	\setlength{\belowcaptionskip}{-1cm}   
	\centerline{\includegraphics[width=7.0cm]{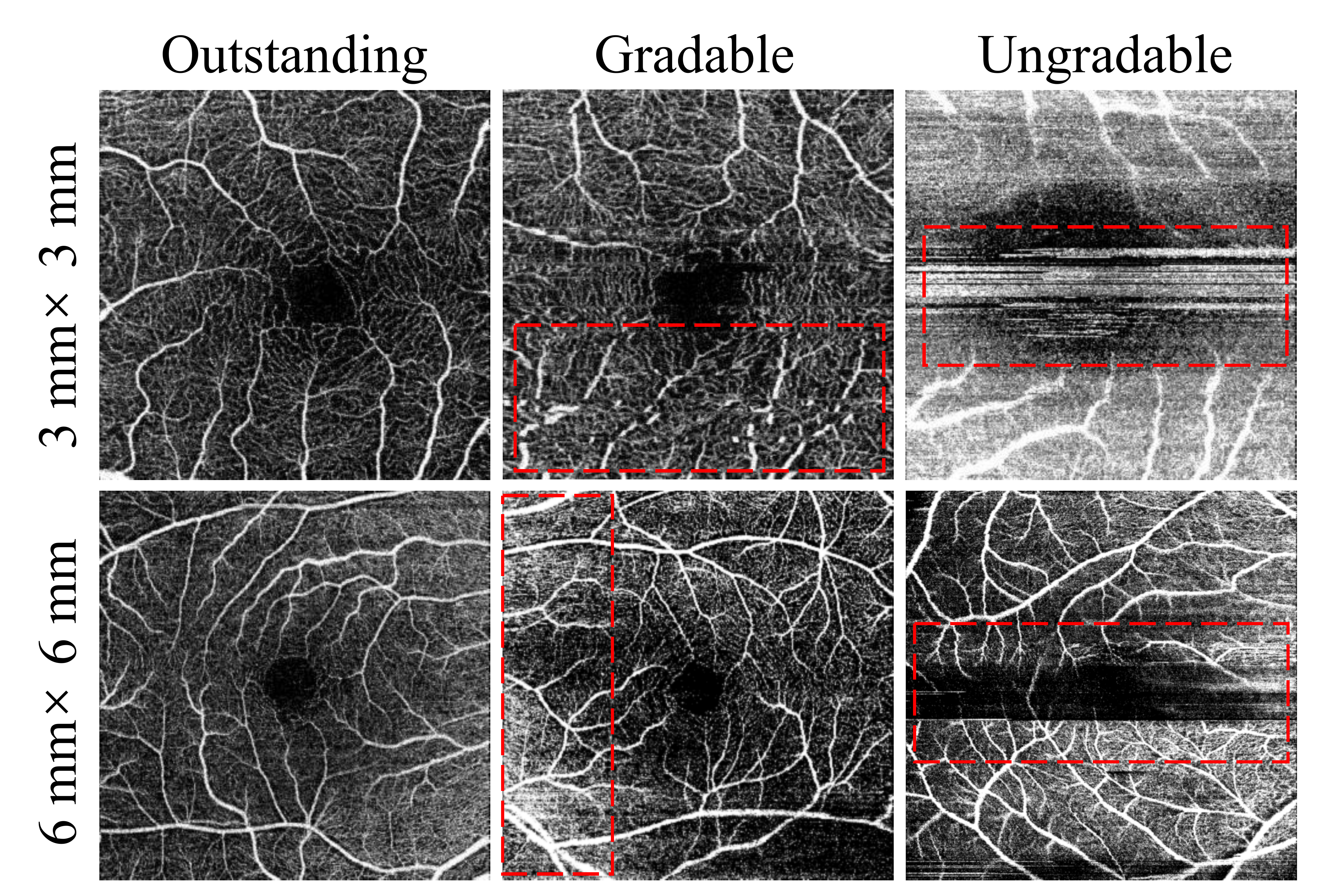}}
	\caption{Representative OCTA examples with different quality grades. Red rectangles highlight low quality regions.}\medskip
	\vspace{-0.7cm}
	\label{intro}
\end{figure} Because of its ability to clearly reveal important anatomical structures such as retinal microvasculature and the avascular zone of the central recess, OCTA has a great potential for accurately diagnosing a variety of fundus-related diseases (e.g. age-related macular degeneration, diabetic retinopathy, etc.) \cite{li2020ipn, peng2022unsupervised}.

OCTA image quality assessment (OIQA) is an important prerequisite in various clinical applications since low-quality images may affect the diagnostic accuracy of both a physician and an intelligent algorithm \cite{cheng2021secret, cheng2021prior, lin2021automated}. OCTA's low-quality issues, such as inadequate illumination, noticeable blur, and low contrast,  may lead to inadequate or even wrong diagnostic decisions. Therefore, automated OIQA is urgently needed.

As a common practice \cite{wang2021deep}, there are three levels for an OCTA image's quality: outstanding, gradable, and ungradable (Figure \ref{intro}). Those three types of images are typically dealt with differently: outstanding images can be directly analyzed or utilized for specific purposes; gradable images can go through image quality enhancement such as denoising, light equalization, contrast enhancement, etc.; ungradable images are considered useless and are often discarded.

There are two main categories in terms of the OIQA solutions: traditional methods and deep learning methods. Traditional methods mainly include distribution-based methods \cite{mittal2012making} and structure-based methods \cite{kohler2013automatic, niemeijer2006image}. The first one constructed a set of quality-related features from a highly regular OCTA scene statistic model, and then fit them with a multivariate Gaussian (MVG) model. The OCTA image quality is then quantified as the distance between two MVG models: one fitted with the features extracted from the test image of interest, and the other fitted with the quality-aware features extracted from the corpus of OCTA images. However, this method typically cannot well measure the distance between distorted and reference images. The second one employs specifically segmented structures like vessels to calculate a global score for noise and blur, and to determine the quality level of the image of interest. However, this kind of methods heavily relies on the accuracy of the structure segmentation and the segmentation task itself may consume a lot of computing resources. Recently, deep learning methods have been widely used for medical image quality assessment \cite{yu2017image, zago2018retinal, fu2019evaluation}. They make use of image features extracted unsupervised-ly or supervised-ly, followed by an image quality classifier or an image quality regressor. A representative deep learning method combines unsupervised features from saliency detection and supervised features from convolutional neural networks (CNNs), and then feeds them to an SVM classifier to identify the quality level. More recently, Wang et al. \cite{wang2022deep} propose to use a ResNet-152 pre-trained by ImageNet to classify OCTA images of different quality levels. Lauermann et al. \cite{lauermann2019standardization} adopt a pre-trained multi-layer deep convolutional neural network to solve the OIQA problem.

Image quality annotations are not always available, especially for emerging modalities. Meanwhile, there may exist many different types of anomalies, exerting difficulty to collecting enough low-quality images that cover a wide range. To address these issues, we propose an unsupervised anomaly-aware framework with test-time clustering for OIQA, namely UNO-QA. UNQ-QA can classify an OCTA image into three grades making use of only a set of outstanding samples during training. Specifically, a neural network with an encoder-decoder structure is trained solely with a set of outstanding samples and then used to discriminate outstanding samples from non-outstanding ones according to their quality scores. Multi-scale feature pyramid pooling is applied to the encoder to extract multi-scale features \cite{gudovskiy2022cflow}. The multi-scale features are concatenated and treated as the anomaly-aware representations, being employed to subdivide non-outstanding into gradable and ungradable. Dimension-reduction is used to reduce the dimension of the anomaly-aware representations, followed by an unsupervised clustering module, so as to subdivide non-outstanding into gradable and ungradable with no supervision involved at all.

The contribution of this work is three-fold:
\begin{itemize}
    \item To the best of our knowledge, UNO-QA is the first unsupervised and hierarchical quality assessment method for ophthalmic images.
    \item UNO-QA novelly extracts and concatenates multi-scale features and combines with feature dimension-reduction and clustering methods, enabling unsupervised quality assessment. Moreover, we perform substantial experiments to identify the optimal combination.
    \item Extensive experiments are conducted on the publicly accessible sOCTA-3×3-10k dataset. UNO-QA outperforms other compared methods, demonstrating the effectiveness of our proposed OIQA framework.
\end{itemize}

\vspace{-0.3cm}
\section{Method}
\label{sec:method}
\vspace{-0.1cm}
\begin{figure*}[htbp]
	\centering
	\vspace{-1.7cm}
	\setlength{\abovecaptionskip}{-0.2cm}   
	\setlength{\belowcaptionskip}{-1cm}   
	\centerline{\includegraphics[height=7.0cm]{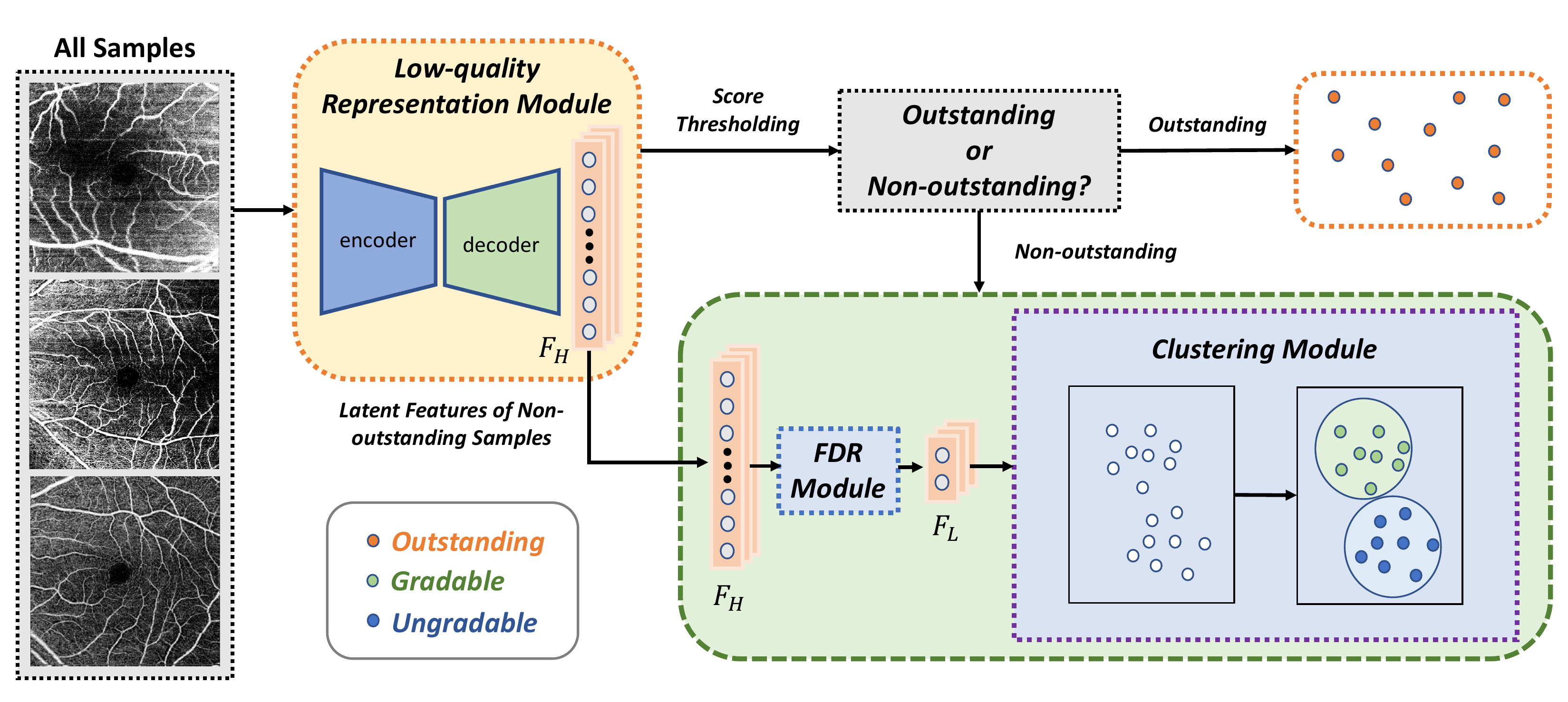}}
	\vspace{-0.3cm}
	\caption{Overview of our proposed framework. In the training stage, we train an encoder with multi-scale pyramid pooling and multiple decoders for multiple scales with only outstanding samples. During the inference stage, all testing samples are fed into the low-quality representation module. Then outstanding samples and non-outstanding samples get separated. For the non-outstanding samples, we extract and concatenate the output features of the decoders and then apply feature dimension-reduction and clustering to subdivide non-outstanding samples into gradable and ungradable samples.}\medskip
	\vspace{-0.5cm}
	\label{model}
\end{figure*}

\subsection{Problem Setting and Notations}
\label{subsec:pre}
For outstanding, gradable, and ungradable samples, we denote them respectively as $\mathcal{I_O}$, $\mathcal{I_G}$, and $\mathcal{I_U}$. Moreover, we define $\mathcal{I_N}$ = \{$\mathcal{I_G}$, $\mathcal{I_U}$\} to represent the set of non-outstanding samples. In the setting of this paper, only a set of high quality samples ($\mathcal{I_O}^{T}$) is available during the training phase, and we aim to automate the tri-classification process of another set of mixed-quality images (\{$\mathcal{I_O}$, $\mathcal{I_G}$, $\mathcal{I_U}$\}) at test time by making use of an anomaly-aware quality representation framework.

\vspace{-0.2cm}
\subsection{Overall Framework}
\label{subsec:ofm}

The overall framework of our proposed UNO-QA is shown in Figure \ref{model}. It consists of three components: (1) A low-quality representation module to distinguish between $\mathcal{I_O}$ and $\mathcal{I_N}$ and to extract the quality-aware representations of $\mathcal{I_N}$; (2) A dimension-reduction module to lower the dimension of the quality-aware representations of $\mathcal{I_N}$; (3) A clustering module to subdivide $\mathcal{I_N}$ into $\mathcal{I_G}$ and $\mathcal{I_U}$.

\vspace{-0.2cm}
\subsection{Low-quality Representation Module}
\label{subsec:adn}

\subsubsection{Anomaly-aware Representation}
\label{subsubsec:ad}
\begin{figure}[htbp]
	\centering
	\setlength{\abovecaptionskip}{-0.2cm}   
	\setlength{\belowcaptionskip}{-1cm}   
	\centerline{\includegraphics[height=3.5cm]{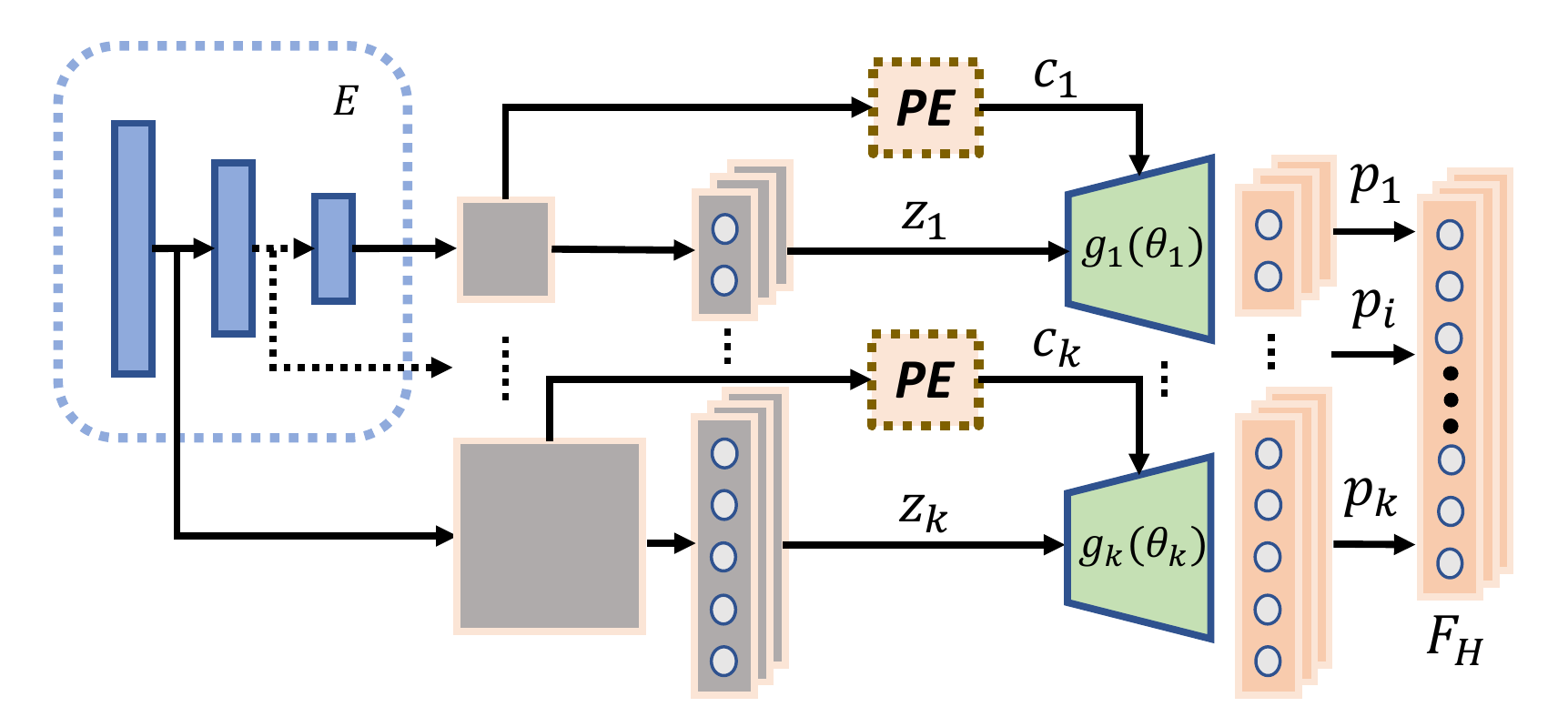}}
	\vspace{-0.3cm}
	\caption{Pipeline of the low-quality representation module.}\medskip
	\vspace{-0.8cm}
	\label{lrm}
\end{figure}

Inspired by \cite{gudovskiy2022cflow}, we design a feature-embedding-based learning framework for the discrimination of $\mathcal{I_O}$ and $\mathcal{I_N}$. As shown in Figure \ref{lrm}, our low-quality representation module has an encoder-decoder structure. The encoder $E$ is pretrained on ImageNet. Since anomalies vary in size and shape, we adopt a multi-scale feature pyramid pooling strategy to provide various receptive fields. Initially, we feed an image of interest into $E$ to obtain feature vectors $\boldsymbol{z}$. Then we employ a MVG $p_{\text{Z}}(\boldsymbol{z})$ as the density function. To represent location features, a conditional vector $\boldsymbol{c}$ which contains spatial location information, is generated using a 2D form of the conventional positional encoding (PE). The decoder $g(\boldsymbol{\theta})$ aims to approximate $p_{\text{Z}}(\boldsymbol{z})$ with an estimated parameterized density $\hat{p}_{\text{Z}}(\boldsymbol{z,c,\theta})$, where $\theta$ is initialized by values sampled from the Gaussian distribution fitted from $\mathcal{I_O}^{T}$. We use the Kullback-Leibler divergence ($D_{KL}$) as the loss function to train the model, namely
\begin{equation}
\begin{aligned}
\label{eq1}
\mathcal{L}({\boldsymbol{\theta})} &= D_{KL} [p_{\text{Z}}(\boldsymbol{z}) || \hat{p}_{\text{Z}}(\boldsymbol{z,c,\theta})].
\end{aligned}
\end{equation}

Since the encoder serves as a multi-scale feature extractor, we need to train $K$ independent decoders $g_{k}(\boldsymbol{\theta}_{k})$. We denote the output of each $g_{k}(\boldsymbol{\theta}_{k})$ as $p_k$ which may be considered as the general spatial semantic feature with a different scale because it jointly incorporates the information of $\boldsymbol{z}$ and $\boldsymbol{c}$.

In the inference phase, since low-quality samples are not observable during the training of the low-quality representation module, the low-quality scores of $\mathcal{I_N}$ should be higher than those of $\mathcal{I_O}$. From the precision-recall curve, we calculate different F1-scores under different thresholds. Then a threshold is obtained by maximizing the F1-score. In this way, we can separate $\mathcal{I_O}$ and $\mathcal{I_N}$ according to their scores.

\vspace{-0.5cm}
\subsubsection{Feature Extraction}
\label{subsubsec:fe}
As mentioned in section \ref{subsubsec:ad}, multi-scale pyramid pooling is employed in the low-quality representation module, which enables the encoder to capture both global and local semantic information with multi-size receptive fields, thus enhancing the low-quality features. For each non-outstanding image, we flatten and concatenate all the multi-scale features $p_1$, $p_2$, ..., $p_k$. We denote the concatenated features of each $\mathcal{I_{N}}^{i}$ as $F_H^i$, where $i$ indexes each non-outstanding image.

\vspace{-0.5cm}
\subsection{Feature Dimension-Reducion}
\label{subsec:fda}
To further divide $\mathcal{I_{N}}$ into $\mathcal{I_{G}}$ and $\mathcal{I_{U}}$, we first reduce the dimension of the representations $F_H$. Since the representations are very complex, the performance of directly employing these high-dimension features $F_H$ to perform clustering may be poor (section \ref{ssec:result}). Therefore, we apply feature dimension-reduction (FDR) to remove noise and some redundant features. We denote the compressed representations as $F_L$. Two FDR methods are considered in this work: Non-negative Matrix Factorization (NMF) and Principal Component Analysis (PCA).
\vspace{-0.2cm}

\subsection{Clustering Module}
\label{subsec:cluster}
Since we only use outstanding samples to train the low-quality representation module, the latent features for images of different quality grades should be dissimilar. Therefore, the representations of $\mathcal{I_G}$ and $\mathcal{I_U}$ shall be different from each other. We thus employ clustering methods like K-means, hierarchical clustering, and Gaussian Mixture Models (GMM) to subdivide $\mathcal{I_N}$ into $\mathcal{I_G}$ and $\mathcal{I_U}$.

\vspace{-0.2cm}

\begin{table*}[ht]
\setlength{\tabcolsep}{1mm}
\newcolumntype{"}{@{\hskip\tabcolsep\vrule width 1.5pt\hskip\tabcolsep}}
\centering
\caption{Comparison results of different pipelines on the two test datasets. LRM stands for low-quality representation module. The best ones are \textbf{bolded} while the second best are \underline{underlined}.}
\vspace{0.5cm}
\label{tab:table1}
\resizebox{1.8\columnwidth}{!}{
\begin{tabular}{ccccccccccccccccccc}
\toprule 
\multicolumn{2}{c}{\multirow{3}{*}{LRM}}                      & \multicolumn{8}{c}{Test} &  \multicolumn{1}{c}{}    & \multicolumn{8}{c}{External Test}    \\ \cline{3-10} \cline{12-19}
& & \multicolumn{2}{c}{K-means}  & \multicolumn{1}{c}{}      & \multicolumn{2}{c}{Hierarchy}                        & \multicolumn{1}{c}{}      & \multicolumn{2}{c}{GMM}         & \multicolumn{1}{c}{} & \multicolumn{2}{c}{K-means}                          &   \multicolumn{1}{c}{} & \multicolumn{2}{c}{Hierarchy}                        &  \multicolumn{1}{c}{} & \multicolumn{2}{c}{GMM}         \\ \cline{3-4} \cline{6-7} \cline{9-10} \cline{12-13} \cline{15-16}  \cline{18-19}
& & \multicolumn{1}{c}{Kappa} & \multicolumn{1}{c}{Acc} &     \multicolumn{1}{c}{}      & \multicolumn{1}{c}{Kappa} & \multicolumn{1}{c}{Acc} &      \multicolumn{1}{c}{}      & \multicolumn{1}{c}{Kappa} & Acc &  \multicolumn{1}{c}{} & \multicolumn{1}{c}{Kappa} & \multicolumn{1}{c}{Acc} & \multicolumn{1}{c}{} & \multicolumn{1}{c}{Kappa} & \multicolumn{1}{c}{Acc} &  \multicolumn{1}{c}{} & \multicolumn{1}{c}{Kappa} & Acc \\ \hline
\multicolumn{1}{c}{\multirow{2}{*}{PaDim}}     & \multicolumn{1}{c}{+PCA} & \multicolumn{1}{c}{13.74}      & \multicolumn{1}{c}{49.27}    &     \multicolumn{1}{c}{}    & \multicolumn{1}{c}{13.69}      & \multicolumn{1}{c}{49.31}    &        \multicolumn{1}{c}{}    & \multicolumn{1}{c}{14.11}      &  
\multicolumn{1}{c}{49.17}    & \multicolumn{1}{c}{}    & \multicolumn{1}{c}{19.50}      & \multicolumn{1}{c}{46.33}    &  \multicolumn{1}{c}{}    & \multicolumn{1}{c}{18.50}      & \multicolumn{1}{c}{45.67}    &  \multicolumn{1}{c}{}    & \multicolumn{1}{c}{18.00}      &      
\multicolumn{1}{c}{45.33}            \\               & \multicolumn{1}{c}{+NMF} & \multicolumn{1}{c}{14.79}      & \multicolumn{1}{c}{50.22}    &    \multicolumn{1}{c}{}    & \multicolumn{1}{c}{14.04}      & \multicolumn{1}{c}{50.15}    &    \multicolumn{1}{c}{}    & \multicolumn{1}{c}{15.08}      &   
\multicolumn{1}{c}{49.98}    &  \multicolumn{1}{c}{}    & \multicolumn{1}{c}{20.50}      & \multicolumn{1}{c}{47.00}    &  \multicolumn{1}{c}{}    & \multicolumn{1}{c}{18.50}      & \multicolumn{1}{c}{45.67}    &  \multicolumn{1}{c}{}    & \multicolumn{1}{c}{19.00}      &  \multicolumn{1}{c}{46.00}           \\ \hline
\multicolumn{1}{c}{\multirow{2}{*}{PatchCore}} & \multicolumn{1}{c}{+PCA} & \multicolumn{1}{c}{33.94}      & \multicolumn{1}{c}{60.34}    &        \multicolumn{1}{c}{}    & \multicolumn{1}{c}{17.28}      & \multicolumn{1}{c}{49.24}    &     \multicolumn{1}{c}{}    & \multicolumn{1}{c}{23.26}      &   
\multicolumn{1}{c}{53.39}    &  \multicolumn{1}{c}{}    & \multicolumn{1}{c}{46.00}      & \multicolumn{1}{c}{64.00}    &  \multicolumn{1}{c}{}    & \multicolumn{1}{c}{45.00}      & \multicolumn{1}{c}{63.33}    &  \multicolumn{1}{c}{}    & \multicolumn{1}{c}{35.00}      &      
\multicolumn{1}{c}{56.67}     \\                      & \multicolumn{1}{c}{+NMF} & \multicolumn{1}{c}{11.33}      & \multicolumn{1}{c}{46.88}    &      \multicolumn{1}{c}{}    & \multicolumn{1}{c}{15.68}      & \multicolumn{1}{c}{49.17}    &        \multicolumn{1}{c}{}    & \multicolumn{1}{c}{13.93}      &  
\multicolumn{1}{c}{47.96}   &  \multicolumn{1}{c}{}    & \multicolumn{1}{c}{24.00}      & \multicolumn{1}{c}{49.33}    &  \multicolumn{1}{c}{}    & \multicolumn{1}{c}{26.50}      & \multicolumn{1}{c}{51.00}    &  \multicolumn{1}{c}{}    & \multicolumn{1}{c}{32.00}      &  \multicolumn{1}{c}{54.67}   \\ \hline
\multicolumn{1}{c}{\multirow{2}{*}{Fastflow}}   & \multicolumn{1}{c}{+PCA} & \multicolumn{1}{c}{42.21}      & \multicolumn{1}{c}{66.98}    &         \multicolumn{1}{c}{}    & \multicolumn{1}{c}{43.78}      & \multicolumn{1}{c}{67.45}    &      \multicolumn{1}{c}{}    & \multicolumn{1}{c}{43.41}      &   
\multicolumn{1}{c}{67.59}    &  \multicolumn{1}{c}{}    & \multicolumn{1}{c}{44.00}      & \multicolumn{1}{c}{62.67}    &  \multicolumn{1}{c}{}    & \multicolumn{1}{c}{31.50}      & \multicolumn{1}{c}{54.33}    &  \multicolumn{1}{c}{}    & \multicolumn{1}{c}{\underline{47.00}}      &      
\multicolumn{1}{c}{\underline{64.67}}     \\                      & \multicolumn{1}{c}{+NMF} & \multicolumn{1}{c}{23.72}      & \multicolumn{1}{c}{56.46}    &        \multicolumn{1}{c}{}    & \multicolumn{1}{c}{22.00}      & \multicolumn{1}{c}{56.42}    &      \multicolumn{1}{c}{}    & \multicolumn{1}{c}{43.18}      &  
\multicolumn{1}{c}{67.42}   &  \multicolumn{1}{c}{}    & \multicolumn{1}{c}{38.50}      & \multicolumn{1}{c}{59.00}    &  \multicolumn{1}{c}{}    & \multicolumn{1}{c}{\underline{47.00}}      & \multicolumn{1}{c}{\underline{64.67}}    &  \multicolumn{1}{c}{}    & \multicolumn{1}{c}{42.50}      &   \multicolumn{1}{c}{61.67}  \\ \hline
\multicolumn{1}{c}{\multirow{2}{*}{Ours}}  & \multicolumn{1}{c}{+PCA} & \multicolumn{1}{c}{46.32}      & \multicolumn{1}{c}{68.06}    &          \multicolumn{1}{c}{}    & \multicolumn{1}{c}{\textbf{53.29}}      & \multicolumn{1}{c}{\textbf{72.61}}    &        \multicolumn{1}{c}{}    & \multicolumn{1}{c}{43.98}      &   
\multicolumn{1}{c}{66.54}    &  \multicolumn{1}{c}{}    & \multicolumn{1}{c}{44.00}      & \multicolumn{1}{c}{62.67}    &  \multicolumn{1}{c}{}    & \multicolumn{1}{c}{44.50}      & \multicolumn{1}{c}{63.00}    &  \multicolumn{1}{c}{}    & \multicolumn{1}{c}{45.00}      &      
\multicolumn{1}{c}{63.33}     \\                      & \multicolumn{1}{c}{+NMF} & \multicolumn{1}{c}{\underline{52.47}}      & \multicolumn{1}{c}{\underline{71.97}}    &       \multicolumn{1}{c}{}    & \multicolumn{1}{c}{50.51}      & \multicolumn{1}{c}{70.69}    &        \multicolumn{1}{c}{}    & \multicolumn{1}{c}{47.90}      &  
\multicolumn{1}{c}{69.27}   &  \multicolumn{1}{c}{}    & \multicolumn{1}{c}{45.50}      & \multicolumn{1}{c}{63.67}    &  \multicolumn{1}{c}{}    & \multicolumn{1}{c}{\textbf{48.50}}      & \multicolumn{1}{c}{\textbf{65.67}}    &  \multicolumn{1}{c}{}    & \multicolumn{1}{c}{45.50}      &  \multicolumn{1}{c}{63.67}   \\ 
\bottomrule 
\end{tabular}
}
\vspace{-0.2cm}
\end{table*}


\begin{table}[htbp]
\centering
\vspace{-0.3cm}
\caption{Ablation studies of our proposed framework. $\mathcal{F}_{QS}$ denotes the quality score. $\mathcal{F}_{single}$ denotes the optimal single-scale features. $\mathcal{F}_{multi}$ denotes the combined multi-scale features. FDR denotes feature dimension-reduction.}
\vspace{0.4cm}
\label{tab:table2}
\begin{tabular}{l|cc}
\toprule 
Metrics                                                   & \multicolumn{1}{c}{Kappa}  & Acc  \\ \hline
$\mathcal{F}_{QS}$  &8.05 &46.78     \\
$\mathcal{F}_{single}$ & 10.34 & 47.72    \\
$\mathcal{F}_{single}$ + FDR & 11.18 & 48.06    \\
$\mathcal{F}_{multi}$ & 50.04 & 70.66    \\
$\mathcal{F}_{multi}$ + FDR (proposed)  & \textbf{53.29} & \textbf{72.61} \\ \bottomrule 
\end{tabular}
\vspace{-0.4cm}
\end{table}

\vspace{-0.1cm}
\section{Experiments}
\label{sec:experiment}
\vspace{-0.1cm}
\subsection{Datasets}
\label{sec:data}
We use the OCTA-25K-IQA-SEG dataset \cite{wang2021deep}, consisting of four subsets. There are two subsets provided for quality assessment, namely sOCTA-3$\times$3-10k and sOCTA-6$\times$6-14k, with a difference in the field of view. In our experiments, we only use the sOCTA-3$\times$3-10k subset. It contains 10,480 $3mm \times 3mm$ superficial vascular layer OCTA (sOCTA) images with three image quality levels (outstanding, gradable, ungradable). Within this dataset, there are 6,915 for training, 2,965 for testing, 300 for validation, and 300 for external testing. The testing set includes 412 outstanding images, 1179 gradable images, and 1374 ungradable images, while the external testing set includes 100 images for each quality grade. In this work, we use all the outstanding samples of the provided training set (961 out of 6915) for training our low-quality representation module. The testing and external testing sets are both used for performance evaluation.
\vspace{-0.3cm}
\subsection{Implementation Details}
\label{ssec:setting}
For the low-quality representation module, the Wide ResNet-50 \cite{zagoruyko2016wide} is employed as the encoder and we implement the decoders following CFLOW-AD \cite{gudovskiy2022cflow}. All compared methods are implemented with the Anomalib library \cite{akcay2022anomalib} and PyTorch Lightning \cite{Falcon_PyTorch_Lightning_2019} using NVIDIA RTX 2080 Ti GPUs. In both training and inference phases, all images are resized to be 320 $\times$ 320. For the clustering module, NMF and PCA are used for feature dimension-reduction. K-means, hierarchical clustering, and GMM are used for clustering. The number of clusters is set as 2 and the maximum iteration index is set as 10000.
\vspace{-0.4cm}

\subsection{Results}
\label{ssec:result}
All methods are evaluated using two metrics, namely Kappa[\%] and Accuracy[\%] (Acc), the results of which are tabulated in Table \ref{tab:table1}. Our low-quality representation module is compatible with most anomaly detection methods. To assess the adaptability of our framework, we analyze different anomaly detection models including PaDim \cite{defard2021padim}, PatchCore \cite{roth2022towards}, and Fastflow \cite{yu2021fastflow}, when incorporated in our UNO-QA pipeline. We observe that our low-quality representation module combined with hierarchy clustering achieves the best classification performance, with PCA and NMF respectively work the best on the testing set and the external testing set when serving as the FDR module.

To further analyze the importance of different components in UNO-QA, we conduct ablation experiments on the testing set and present the results in Table \ref{tab:table2}. In all ablation experiments, PCA and hierarchical clustering are respectively adopted in the FDR module and the clustering module since their combination works relatively the best (Table \ref{tab:table1}). We observe that the latent features extracted from the low-quality representation module are better than the quality scores and identify the importance of the multi-scale pyramid pooling operation. Moreover, employing FDR for either single-scale features or concatenated multi-scale features makes a difference, which clearly suggests the importance of dimension reduction before clustering.

\vspace{-0.3cm}
\section{Conclusion}
\label{sec:conclusion}
\vspace{-0.2cm}
In this paper, we propose a novel framework for unsupervised and hierarchical (two-level) OIQA. We distinguish between outstanding and non-outstanding based on a feature-embedding based low-quality representation module and then extract multi-scale features from the low-quality representation module to perform feature dimension-reduction and clustering. To the best of our knowledge, our method is the first one to apply unsupervised learning to three-level OCTA image quality assessment tasks, and it has a great potential to be extended to other types of medical images.

\vspace{-0.4cm}
\section{Acknowledgments}
\vspace{-0.2cm}
This study was supported by the Shenzhen Basic Research Program (JCYJ20190809120205578); the National Natural Science Foundation of China (62071210); the Shenzhen Science and Technology Program (RCYX20210609103056042); the Shenzhen Basic Research Program (JCYJ2020092515384\\7004); the Shenzhen Science and Technology Innovation Committee Program (KCXFZ2020122117340001).
\vspace{-0.2cm}



\small
\bibliographystyle{IEEEbib}

\end{document}